# From Explainability to Action: A Generative Operational Framework for Integrating XAI in Clinical Mental Health Screening

Ratna Kandala[1]
[1]Independent Researcher
Lawrence, KS, USA
ratnanirupama@gmail.com

Akshata Kishore Moharir[2]
[2]Independent Researcher
Oregon, Sherwood, USA
akshatankishore5@gmail.com

Divya Arvinda Nayak[3]
[3]Independent Researcher
Arlington, Virginia, USA
nayakdivya01@gmail.com

*Abstract*—Explainable Artificial Intelligence (XAI) has been presented as the critical component for unlocking the potential of machine learning in mental health screening (MHS). However, a persistent "lab-to-clinic" gap remains. Current XAI techniques, such as SHAP and LIME, excel at producing technically faithful outputs such as feature importance scores, but fail to deliver clinically relevant, actionable insights that can be used by clinicians or understood by patients. This disconnect between technical transparency and human utility is the primary barrier to real-world adoption.

This paper argues that this gap is a translation problem and proposes the *Generative Operational Framework*, a novel system architecture that leverages Large Language Models (LLMs) as a central translation engine. This framework is designed to ingest the raw, technical outputs from diverse XAI tools and synthesize them with clinical guidelines (via RAG) to automatically generate human-readable, evidence-backed clinical narratives.

To justify our solution, we provide a systematic analysis of the components it integrates, tracing the evolution from intrinsic models to generative XAI. We demonstrate how this framework directly addresses key operational barriers, including workflow integration, bias mitigation, and stakeholder-specific communication. This paper also provides a strategic roadmap for moving the field beyond the generation of isolated data points toward the delivery of integrated, actionable, and trustworthy AI in clinical practice.

## I. INTRODUCTION

The world is facing a significant mental health crisis [1], and in response, Artificial Intelligence (AI) and Natural Language Processing (NLP) have shown immense promise, powering a new generation of mental health screening (MHS) tools capable of analyzing subtle patterns in text, speech, and electronic health records (EHRs) [2]–[4]. Yet, this potential is almost entirely blocked by a single, formidable barrier: opacity. The most powerful models are "black boxes," whose internal logic is inscrutable [5] and unacceptable in high-stakes domains such as mental health [2]. Clinicians cannot take professional responsibility for an AI's diagnoses and recommendations without knowing the reasoning behind the output, and patients will not trust diagnoses from an algorithm they cannot understand [6]–[8].

Explainable AI (XAI) emerged as the logical solution to this crisis of trust [9]. An entire field of techniques - most notably LIME and SHAP - succeeded in cracking open the black box, allowing us to see which features a model deemed important [2]. However, this technical victory exposed a deeper, more practical failure: we have progressed toward solving the issue of 'technical transparency' but not 'human utility.' For example, a SHAP plot given to a psychiatrist in a 15-minute consultation is not an explanation; it is just more data to interpret. This has only created a persistent "algorithm-to-clinic" gap [5], [10]–[12], leaving the field with an abundance of XAI techniques but an almost total lack of clinically integrated XAI systems.

To address this, this paper argues that this critical gap is a translation problem and proposes a solution: the Generative Operational Framework, a novel system architecture designed specifically to translate the technical outputs of XAI into clinically fluent, actionable narratives. The core of our proposal is the use of an LLM as an integrated "translation engine." Instead of presenting a clinician with a raw feature-importance list, our proposed framework feeds those technical outputs directly to a generative model. This model, augmented by Retrieval Augmented Generation (RAG) from established clinical guidelines (e.g., DSM-5) and patient-specific history, synthesizes all available information into a concise, human-readable summary. This output directly answers the clinician's need: "What did the AI find, what is its reasoning process, what evidence supports its reasoning, and what is its level of confidence?" In this regard, Figure 1 illustrates how Explainable AI (XAI) is central to the mental health screening lifecycle.

To build the case for this framework, the rest of the paper provides the necessary context and justification. We conduct a systematic survey of the XAI components our framework synthesizes, analyzing their evolution not as a simple history, but as a series of technical advancements that make our proposal both possible and necessary in sections II and III. We then provide a comparative analysis of these tools (section IV), critically evaluate the operational pathways our framework is built to navigate (section V), and highlight the benchmarking

(section VI) and ethical challenges that any such system must overcome (section VII).

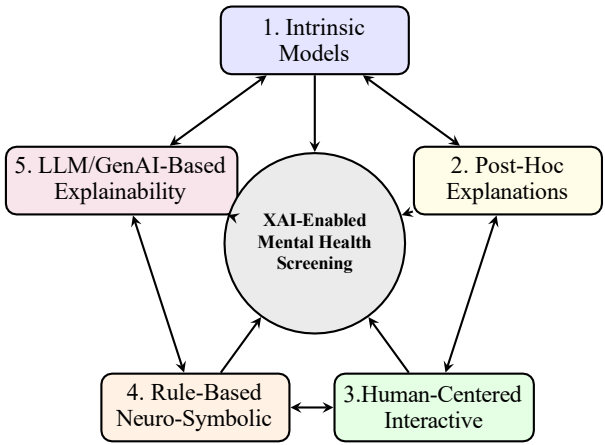

Fig. 1: XAI at the Core of Mental Health Screening

## II. THE EVOLUTION OF EXPLAINABILITY: A TECHNICAL JOURNEY

The quest for explainability in MHS was not a single discovery but an evolution driven by the needs of its stakeholders - patients, clinicians, and regulators [13], [14]. For patients, explanations are crucial to foster trust and engagement; for clinicians, they must support professional judgment and shared decision-making; and for regulators, they must enable auditability and ensure alignment with ethical frameworks like the US's Health Insurance Portability and Accountability Act (HIPAA) and the EU's General Data Protection Regulation (GDPR). This journey progressed from demanding simple transparency to requiring sophisticated, human-readable narratives. The methods used followed this path, evolving across five major stages (Figure 1):

**(a) The Foundation: Intrinsic Models (Explainability by Design)**

The first and most direct path to explainability was to not create a black box in the first stage. This "white-box" approach favors models whose internal logic is directly understandable, such as:

- *Logistic Regression and Decision Trees:* Foundational models like these remain in active use, not just as simple tools but as crucial, transparent baselines. A recent study by [15] found logistic regression to be the most balanced and interpretable model for a clinical support task, achieving 97.3% accuracy with a low false-negative rate. They provide a vital benchmark against which the performance-interpretability trade-off of more complex models must always be justified [2], [16].

**(b) The Compromise: Peering Inside with Post-Hoc Techniques**

As the data mining challenge moved to unstructured text and multimodal data in the second stage, simple intrinsic models were no longer sufficient. The field needed the predictive power of deep learning, forcing a compromise: if we must use a black box, we must also build a tool to look inside it. This led to post-hoc methods, which interpret a model *after* a prediction is made. Local Interpretable Model-agnostic Explanations (LIME) and SHapley Additive exPlanations (SHAP) became the standard tools for analyzing deep learning models in MHS research [2], [5].

**(c) The Realization: Moving from Data to Dialogue** A crucial realization soon followed in the third stage: a SHAP plot or a list of LIME keywords is not a clinical explanation; rather, it is raw data that still requires expert interpretation. This gap sparked the move toward human-centered paradigms that align with clinical reasoning.

*Visual, Interactive, and Case-based:* Explanation evolved from a static output into an interactive process. Visual analytics dashboards allowed clinicians to probe a model's reasoning directly [17], [18]. This was complemented by Case-Based Reasoning (CBR), which explains a new prediction by finding similar past cases—a method highly intuitive to clinical training [19]. A 2024 framework, ProtoDep [20], perfectly captures this shift, using prototype learning to explain a user's depression risk by comparing their social media profile to representative examples of depressive behavior.

**(d) The Hybrid: Combining Logic and Learning**
Seeking to blend the rigor of intrinsic rules with the power of neural networks, neuro-symbolic systems emerged in the fourth stage. These systems encode structured knowledge, such as DSM-5 criteria, into formal logic that can guide or constrain a neural network [21], [22]. This hybrid approach ensures that even complex pattern recognition adheres to established clinical knowledge, making the system's outputs inherently auditable.

**(e) The New Vanguard: LLM-Generated Explanations**
The most recent and transformative (fifth) stage is the use of LLMs as the explainability engine itself. This paradigm shift goes beyond extracting explanations to *generating* them in a human-understandable, context-aware language [23], [24].
**Chain-of-Thought (CoT) Prompting:** CoT forces an LLM to "show its work" by generating a step-by-step reasoning chain before giving a final answer. In MHS, this mimics a clinical diagnostic workflow - moving from sentiment analysis to causal reasoning - making the model's logical path transparent by design [25], [26].
**Retrieval-Augmented Generation (RAG):** RAG connects the LLM's reasoning to a verifiable body of external knowledge. This is a critical safety and explainability feature for clinical use. A RAG-based system can retrieve relevant passages from EHRs or clinical guidelines (e.g., DSM-5) and *cite them as direct evidence* for its recommendations. Frameworks like MedRAG and RAGnosis are pioneering this approach, delivering not just answers, but traceable, evidence-backed reasoning from unstructured clinical text [27], [28].

These technical developments not only give us the necessary tools but also establish a framework for investigating how these methods function in practical mental health screening applications. Now that we have explored the advancement of explainability approaches, the following section will focus on their application to particular challenges.

## III. XAI IN THE WILD: FROM SOCIAL MEDIA SIGNALS TO CLINICAL INSIGHTS

The true test of the XAI evolution is how these techniques are applied to solve concrete clinical problems. The progression from post-hoc analysis to generative narratives becomes clear when mapping these tools to specific MHS data mining challenges, as elaborated below:

### A. Challenge 1: Mining Depression from Unstructured Social Media

Mining public social media for signs of depression presents a classic data mining problem: the data is noisy, vast, and unstructured, requiring powerful models. This domain provides a perfect microcosm of XAI's evolution.

**Act 1: Post-Hoc Analysis.** Initial approaches applied powerful transformers like BiLSTM and MentalBERT to user posts [29], [30] where these models achieved high accuracy but were opaque. Researchers applied LIME and AGRAD to peer inside, revealing predictive word categories (e.g., "hopeless," "alone") [2]. While an important first step, this highlighted a critical gap: knowing *which* words were important is not the same as understanding the *clinical reasoning*.

**Act 2: Case-Based Reasoning.** The field moved to create more intuitive explanations. Frameworks like ProtoDep [20] answered a more clinically relevant question: not "what words did the model see?" but "who does this patient look like?" By using prototype learning to compare a user's profile to representative examples of depressive behavior, the explanation became a comparison, a method far more aligned with clinical practice.

**Act 3: Generative Assessment.** Most recently, LLMs have begun to bridge the gap between informal social media expression and formal clinical instruments. Research has shown that a model like Llama-2 can not only automatically complete a clinical standard like Beck's Depression Inventory (BDI) from a user's posts but can also provide a natural language rationale for its classification, effectively automating the assessment and documentation process [31].

### B. Challenge 2: Assessing Suicide Risk from Clinical Notes

In the high-stakes domain of suicide risk assessment from EHR clinical notes, explainability is a non-negotiable safety requirement [3]. This field demonstrates the powerful synthesis of post-hoc methods with generative AI to create clinically actionable tools. A state-of-the-art approach by [32] used an XGBoost model to predict risk from notes. However, rather than simply presenting a SHAP plot, they built a pipeline that feeds the SHAP outputs (the key risk and protective factors) into an LLM. The LLM's task was to translate this raw data into a clinically coherent narrative aligned with the formal SAFE-T assessment framework. This represents a crucial evolutionary leap: from explaining *what words mattered* to the model, to explaining the *clinical significance* of the prediction in the language of the clinician.

### C. Challenge 3: Detecting Anxiety and Stress from Multimodal Data

Mental health is not just expressed in text. New frontiers involve mining multimodal data, including video and speech. Deep learning systems are being developed to detect human stress from video data, opening the door to noninvasive, continuous monitoring [33]. Simultaneously, models like GPT-4 are proving highly adept at inferring social anxiety symptom strength directly from the transcripts of clinical interviews, correlating strongly with patient self-reports [34]. In this domain, XAI is essential for identifying which multimodal cues (e.g., vocal tone, specific pauses, facial expressions) are driving the model's assessment, allowing clinicians to validate the automated findings against their own holistic view of the patient.

Having looked at the application of explainability to certain challenges, the next section will list out the key metrics for evaluating clinical explanations.

## IV. CHOOSING THE RIGHT LENS: A COMPARATIVE AND EVALUATION FRAMEWORK

This evolutionary journey has produced a diverse toolkit, but it has also made the path to implementation more complex. Choosing the right XAI method is not a technical decision alone; it is a clinical and ethical one, driven by a series of trade-offs. To navigate these choices, we must first define what makes an explanation "good" and understand the critical lack of standardized data to measure it.

### A. Key Metrics for Evaluating Clinical Explanations

Judging an explanation requires moving beyond computational accuracy to a human-centered framework [35], [36]. Any evaluation must measure:

- **Fidelity:** Does the explanation accurately reflect the model's true internal logic? [35]
- **Comprehensibility:** Can the intended user (clinician or patient) actually understand the explanation? [37]
- **Clinical Utility:** Does the explanation provide actionable insights that lead to better diagnostic decisions? [5]
- **Calibrated Trust:** Does the explanation help the user build an appropriate level of confidence, avoiding both over-reliance and baseless skepticism? [38], [39]
- **Timeliness:** Can the explanation be generated and understood within the time constraints of a clinical consultation? [40]

*Critical Gap:* While these dimensions are known, they are rarely evaluated together in rigorous clinical validation studies [5]. This gap between metric development and real-world testing remains a primary barrier to translation.

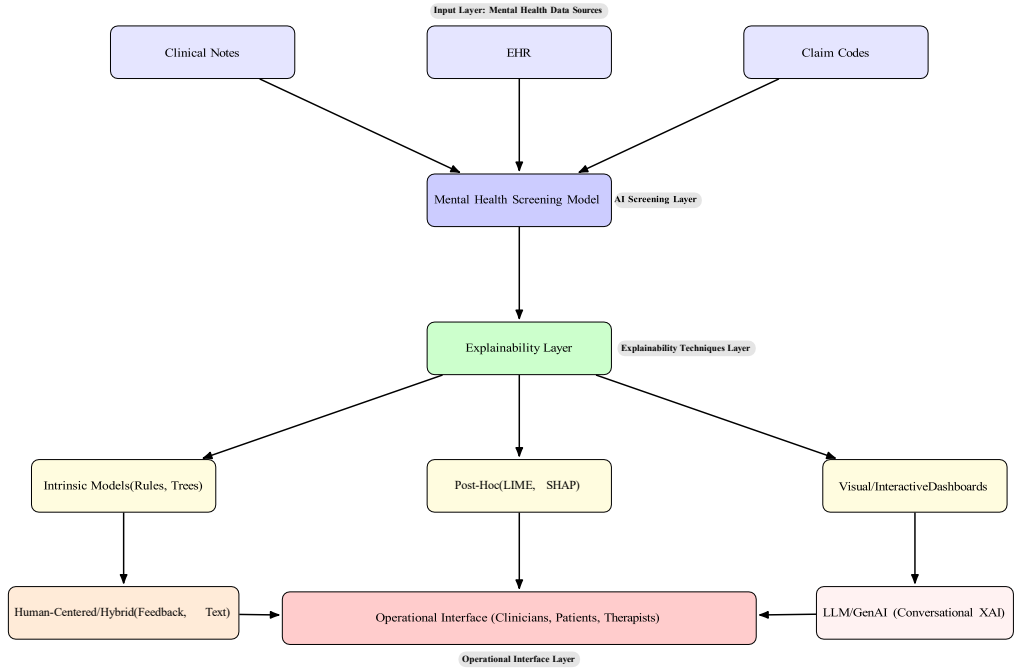

Fig. 2: Conceptual Framework: Explainable AI Approaches for Mental Health Screening

We now turn to a critical obstacle that hampers the practical deployment of XAI techniques: the absence of adequate benchmarking standards and comprehensive datasets.

## V. Benchmarking and the Critical Data Void

This evaluation gap is exacerbated by a lack of standardized data. Clinical translation is impossible without robust benchmarks, yet the current landscape is fragmented.

**Available Datasets:** Tools like **DAIC-WOZ** (clinical interviews), **CLPsych Shared Tasks** (social media), **AVEC Challenges** (multimodal emotion), and **MIMIC-IV** (EHR data) have all pushed the field forward [41]–[45].

**Their Limitations:** Each dataset carries limitations. DAIC-WOZ lacks demographic diversity; CLPsych is biased against non-online populations; AVEC focuses on emotion, not diagnostics; and MIMIC-IV is focused on critical care, not outpatient mental health. Most importantly, no current dataset was built to simultaneously evaluate **clinical fidelity** (DSM-5 alignment), **explanation utility** (actionability), **demographic fairness**, and **temporal stability**. This void makes direct comparison of new XAI techniques nearly impossible and directly impedes regulatory approval.

### A. Navigating the Trade-Offs

Given this landscape, selection involves balancing competing virtues (Table I). Intrinsic methods (rules, trees) offer perfect transparency and speed, making them ideal for high-stakes regulatory audits or triage. However, they lack the predictive power for complex data. Post-hoc methods (LIME, SHAP) and visual dashboards provide a flexible balance, allowing the use of powerful black-box models while offering crucial insights [46]–[48].

The newest LLM/GenAI approaches offer the highest potential for user engagement and personalized, conversational explanations. They excel at shared decision-making but introduce new and serious risks of hallucination, faithfulness, and regulatory compliance [49], [50]. This entire landscape of choice is visualized in the decision tree in Figure 4.

To advance XAI from research to clinical practice, there need to be clear operational pathways that take into account technical, human, and organizational factors. This section illustrated the path from the algorithm to the bedside, as well as points out significant decisions points where XAI implementations either perform well or fail to function.

With an understanding of the available tools, their inherent trade-offs, and the data challenges that constrain them, we now shift focus to the practical question of implementation: how can XAI systems be successfully translated from research prototypes into functioning clinical tools?

## VI. From the Lab to the Clinic: Operationalizing the XAI Pipeline

Knowing the tools and their trade-offs is only half the story. The ultimate goal is clinical translation, a journey from the algorithm to the bedside that is fraught with technical, human, and organizational hurdles. Successfully operationalizing XAI requires a clear pathway that integrates these new tools into existing, complex clinical ecosystems (Figure 5).

### A. Workflow Integration Stages

**Stage 1 - Pre-deployment:** This stage is defined by needs assessment and co-design. Early and continuous involvement of clinicians to map existing workflows and identify pain points is the single most critical factor for adoption [56].

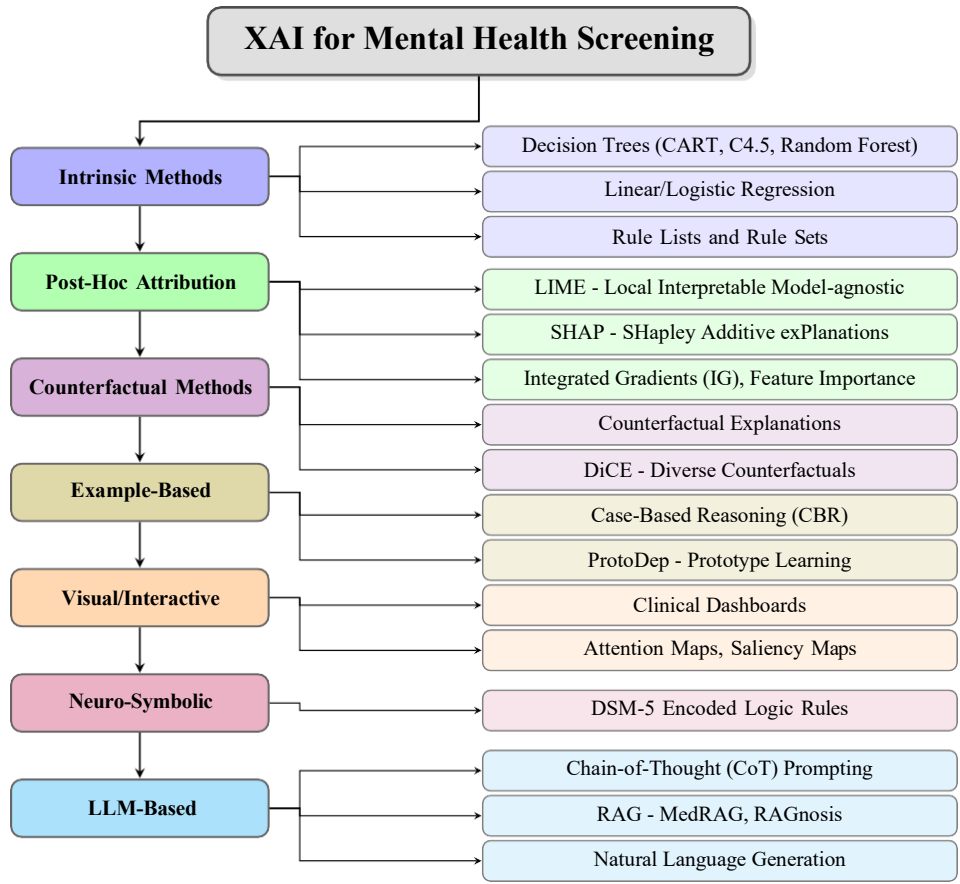

Fig. 3: Taxonomy of XAI techniques for Mental Health Screening

**Stage 2 - Pilot testing:** Initial validation occurs in controlled clinical settings. Here, key indicators are measured: Does the explanation add to the cognitive load? How does it affect decision time and physician confidence? [70].

**Stage 3 - Full integration:** This involves deep integration of the EHR, staff training, and continuous monitoring. The key success factor is an explanation that fits "ambiently" into the consultation without demanding a disruptive new process [71].

### B. Stakeholder-Specific Pathways

An explanation is only useful if it serves the needs of its specific audience. A "one-size-fits-all" explanation will fail.

**For Clinicians:** The output must be actionable within the tight time constraint of a patient visit. A concise visual dashboard or a single-paragraph narrative summary (such as the LLM-SHAP pipeline) is often more effective than a detailed, raw feature importance graph [72].

**For Patients:** Explanations must be translated from technical jargon into empathetic, comprehensible, and clinically valid language to support shared decision-making and build trust [73].

**For Administrators:** The system must provide clear audit trails and documentation demonstrating regulatory compliance (e.g., with HIPAA or GDPR) to justify institutional adoption [49].

### C. Implementation Barriers and Mitigation

The path is blocked by significant barriers. **Technically**, EHR interoperability remains the main bottleneck [74]. **Humanly**, trust is not granted automatically; it must be built iteratively over time [75]. **Organizationally**, the lack of clear accountability frameworks who is responsible if an explained AI makes a mistake? slows adoption [76].

The *critical path* to success is therefore not a one-time "big bang" deployment, but an iterative rollout centered on continuous feedback loops with all stakeholders [71], as elaborated in the next section.

## VII. THE PATH AHEAD: UNRESOLVED CHALLENGES AND FUTURE QUESTS

Our story of XAI in MHS has no ending; it simply arrives at the next frontier. Successfully deploying these advanced tools requires resolving profound technical, human, and infrastructure challenges. These gaps define the critical path forward for the data mining and clinical research communities.

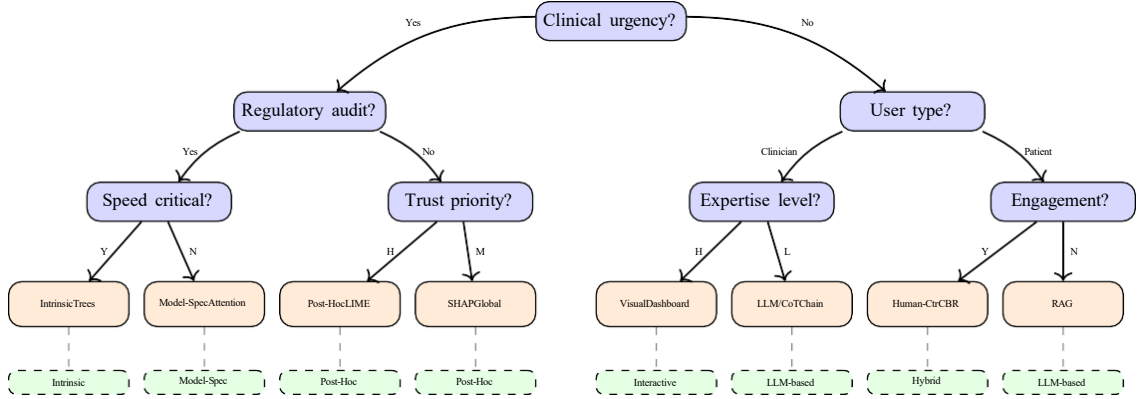

Fig. 4: Decision tree for XAI method selection in mental health screening where Y=Yes, N=No, H=High, M=Moderate, L=Low

TABLE I: Comparative strengths and weaknesses of major XAI classes for mental health screening

| XAI Class | Interpretability | Speed/Cost | User/Clinician Trust | Context Fit (Clinical Usability) |
| --- | --- | --- | --- | --- |
| Intrinsic Models (Rules, Trees, Regression) | Very high; logic is explicit | Very fast | Very high; trusted in clinic and audit [51], [52] | Excellent for triage, intake, audit, regulatory apps; less capable with nonlinear/high-dim data |
| Post-Hoc (LIME, SHAP, Counterfactuals) | Moderate–high per-case [46], [48] | Moderate; can be slow with complex models | Moderate; trusted if concise, but explanations may be technical [53], [54] | Flexible for "black box" models; less optimal for urgent clinical flows |
| Visual/Interactive (Dashboards, Overlays) | High (summary/trends), moderate (detail) | Fast; real-time display possible | High with UX alignment; supports shared understanding [55], [56] | Best for ongoing monitoring, self-help, population analytics |
| Human-Centered/Hybrid (Text, Example, Feedback) | Very high (personalized, lay language) | Fast; text generation is quick | Very high; supports engagement, learning [53], [57] | Ideal for therapy apps, education, patient engagement; faithfulness critical |
| Model-Agnostic (LIME, SHAP, Counterfactuals) | High per-case, moderate globally [46], [48] | Moderate to slow | Good for experts, but non-expert users need tailored output [53] | Used for research, auditing, or when model flexibility is needed |
| Model-Specific (Attention, Feature Importances) | Moderate–high (if model known) | Fast (built-in) | Moderate; depends on user familiarity [58], [59] | Suitable for researchers, advanced clinics; less generalizable for laypersons |
| LLM/GenAI-Based XAI | High—rich, conversational, scenario, CoT [23], [49], [60] | Moderate; scales with LLM size/prompt | High if justified, context-aware explanations; but faithfulness a challenge [61] | Emerging for triage, education, digital therapy, self-care; needs oversight for clinical use |

### A. Critical Technical Challenges

**Explanation Faithfulness: Challenge:** LLM-generated explanations, although articulate, may not accurately represent the model's true reasoning, creating plausible but clinically misleading justifications [49], [77]. **Opportunity:** Develop automated faithfulness audits and new model architectures that intrinsically link generative reasoning to predictive output.

**Privacy-Preservation: Challenge:** Explanations, especially those based on cases or generative ones, can accidentally leak sensitive patient data from their training set, violating privacy [74], [75]. **Opportunity:** Advance privacy-preserving XAI frameworks using differential privacy and federated learn-

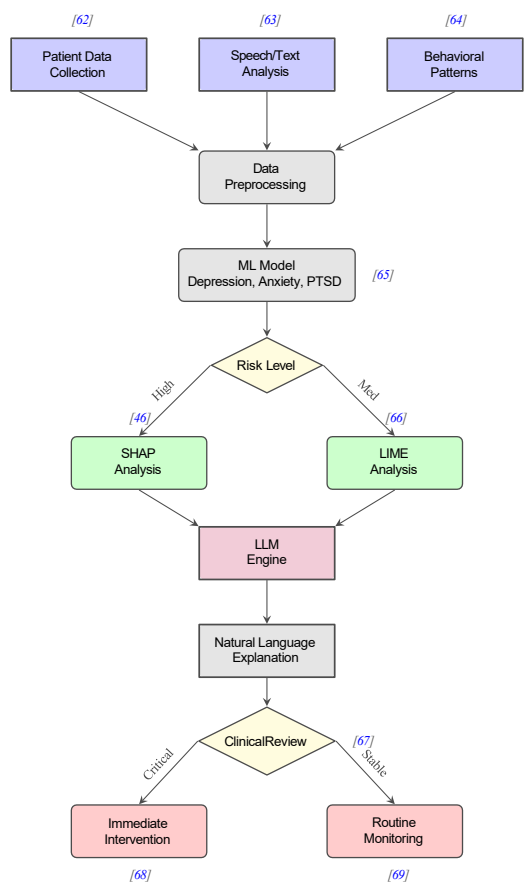

Fig. 5: Architecture of an explainable AI system for mental health screening, integrating multimodal data collection (patient records, speech analysis, and behavioral patterns), machine learning-based risk assessment, post-hoc explanation methods (SHAP and LIME), and LLM-powered natural language generation to facilitate clinical decision-making and intervention planning

ing to generate explanations without exposing raw patient data.

**LLM Hallucination: Challenge:** GenAI systems may "hallucinate" or fabricate clinical justifications that are not evidence-based, a critical safety failure [49], [77]. **Opportunity:** Mandate and refine RAG implementations that anchor all generative explanations to a verifiable corpus of clinical guidelines or patient-specific EHR data.

### B. Human-Centered Challenges

**Context Adaptation: Challenge:** A generic explanation fails diverse populations that vary by age, health literacy, and clinical need [71], [75]. **Opportunity:** Design adaptive XAI systems that dynamically adjust the complexity and modality (visual, text, conversational) of an explanation based on the end-user's profile and needs.

**Bias and Fairness: Challenge:** If a model is biased, its explanation will only justify a biased result. Bias can exist in both the prediction and the explanation itself, particularly for marginalized groups [75], [78]. **Opportunity:** Mandate fairness audits not just for model predictions, but for explanation utility across all demographic groups *before* deployment.

**Stakeholder Engagement: Challenge:** Tools built without end-users become clinically irrelevant "shelfware" [71]. **Opportunity:** Establish permanent, participatory design frameworks that maintain continuous feedback loops between researchers, clinicians, and patients.

### C. Infrastructure and Evaluation Gaps

**Benchmarking Void: Challenge:** As established, there are no standardized datasets or metrics to assess the longitudinal clinical utility of XAI in MHS [41], [49], [56]. **Opportunity:** A grand challenge for the community is the creation of open, multimodal, and demographically diverse benchmarks that specifically measure the utility and reliability of the explanation over time.

**Interoperability: Challenge:** Most XAI modules are standalone prototypes that cannot communicate with EHRs, telehealth platforms, and mobile apps where they are needed [56], [74]. **Opportunity:** Develop universal APIs and adhere to data standards such as FHIR to ensure XAI modules can function as interoperable "plugins" within any clinical system.

## VIII. CONCLUSION: THE NEXT CHAPTER OF THE STORY

The story of XAI in mental health screening is one of rapid and necessary evolution, a journey from opaque black boxes to generative clinical narratives. We have traced this path from the simple transparency of intrinsic models, through the necessary compromise of post-hoc methods like SHAP and LIME, to the crucial realization that clinicians and patients need human-centered insights, not just raw data. The arrival of LLMs has opened the final act, where AI can generate its own evidence-backed reasoning.

However, this story is unfinished. The path forward demands that the research community move beyond technical demonstrations to address the difficult problems of clinical validation, ethical integrity, and operational integration. Future work must focus on (a) adaptive contextual explanations for diverse users; (b) longitudinal studies demonstrating real-world impact on patient outcomes; (c) robust privacy-preserving frameworks; and (d) the creation of interoperable benchmarks to fairly measure our progress. The goal of the data mining community is to ensure that the next generation of AI tools is not just intelligent but comprehensible, ethical, and fully aligned with the values and needs of mental health care.